\documentclass[11pt,twocolumn,letterpaper]{article}
\usepackage{graphicx}
\usepackage{times}
\usepackage{amsmath}
\usepackage{amssymb}
\usepackage{url}

\usepackage{ctable}
\usepackage{multirow}
\usepackage{stfloats}
\usepackage{array}
\usepackage{natbib}
\setcitestyle{authoryear,open={[},close={]}}

\usepackage{gensymb}
\usepackage{balance}
\usepackage{float}

\usepackage{caption}
\usepackage{subcaption}

\newcolumntype{Q}{>{\centering\arraybackslash}m{1cm}}
\newcolumntype{O}{>{\centering\arraybackslash}m{0.8cm}}
\newcolumntype{P}{>{\centering\arraybackslash}m{1.4cm}}
\newcolumntype{L}{>{\centering\arraybackslash}m{1.5cm}}
\newcolumntype{S}{>{\raggedright\arraybackslash}m{3.8cm}}
\newcolumntype{K}{>{\raggedright\arraybackslash}m{2.3cm}}
\newcolumntype{A}{>{\centering\arraybackslash}m{2cm}}
\newcolumntype{R}{>{\centering\arraybackslash}m{0.8cm}}
\newcolumntype{V}{>{\centering\arraybackslash}m{1.6cm}}

\date{}
\begin{document}

\title{Unconstrained Matching of 2D and 3D Descriptors\\ for 6-DOF Pose Estimation
}
\author{Uzair Nadeem$^1$, Mohammed Bennamoun$^1$, Roberto Togneri$^2$, Ferdous Sohel$^3$\\
1 Department of Computer Science and Software Engineering,\\ The University of Western Australia\\
2 Department of Electrical, Electronics and Computer Engineering,\\ The University of Western Australia\\
3 College of Science, Health, Engineering and Education, Murdoch University\\
{\tt\small \{uzair.nadeem@research.,mohammed.bennamoun@,roberto.togneri@\}}\\{\tt\small uwa.edu.au,f.sohel@murdoch.edu.au}
}




\maketitle

\begin{abstract}
This paper proposes a novel concept to directly match feature descriptors extracted from 2D images with feature descriptors extracted from 3D point clouds. We use this concept to directly localize images in a 3D point cloud. We generate a dataset of matching 2D and 3D points and their corresponding feature descriptors, which is used to learn a Descriptor-Matcher classifier. To localize the pose of an image at test time, we extract keypoints and feature descriptors from the query image. The trained Descriptor-Matcher is then used to match the features from the image and the point cloud. The locations of the matched features are used in a robust pose estimation algorithm to predict the location and orientation of the query image. 
We carried out an extensive evaluation of the proposed method for indoor and outdoor scenarios and with different types of point clouds to verify the feasibility of our approach. 
Experimental results demonstrate that direct matching of feature descriptors from images and point clouds is not only a viable idea but can also be reliably used to estimate the 6-DOF poses of query cameras in any type of 3D point cloud in an unconstrained manner with high precision. 
\vspace{1cm}


\end{abstract}

\begin{figure*}
\begin{subfigure}{.33\textwidth}
\centering
\includegraphics[width=5.5cm]{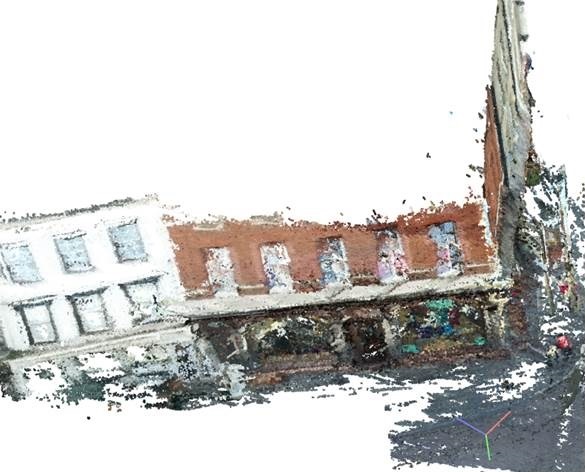}
\caption{3D point cloud}
\label{fig:1a}
\end{subfigure}\hfill
\begin{subfigure}{.33\textwidth}
\centering
\includegraphics[width=5.5cm]{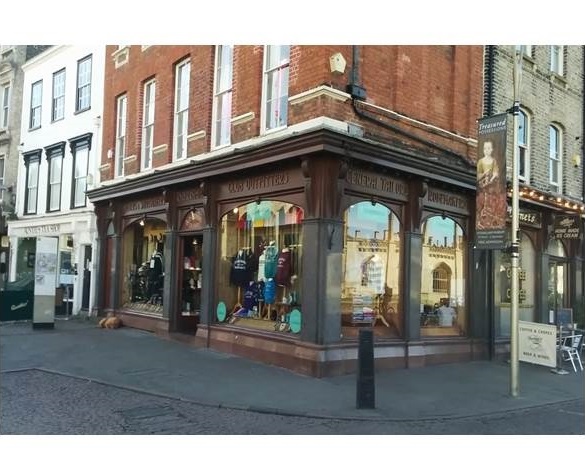}
\caption{2D query image}
\label{fig:1b}
\end{subfigure}\hfill
\begin{subfigure}{.33\textwidth}
\centering
\includegraphics[width=5.5cm]{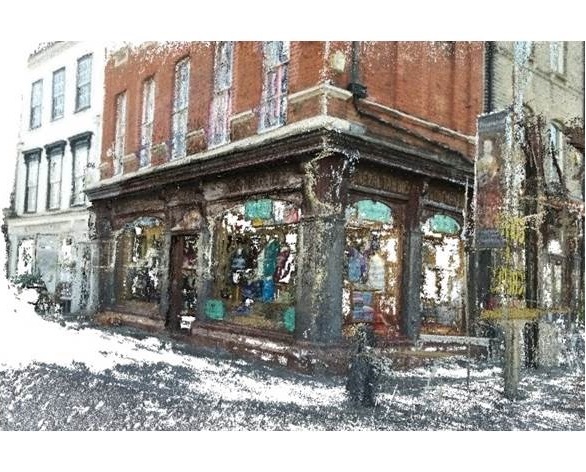}
\caption{Localization results of (b) in (a)}
\label{fig:1c}
\end{subfigure}
\caption{(a) A section of the 3D point cloud from Shop Facade dataset \citep{posenet}. (b) An RGB query image to be localized in the 3D point cloud (c) Visualization of the area of the 3D point cloud, identified by our technique as the location of the query image.}
\label{fig:result}
\end{figure*}

\section{Introduction}

The numerous applications of 3D vision, such as augmented reality, autonomous driving and robotic locomotion, aided by the ever increasing computational power of modern machines have caused a steep increase of interest in 3D vision and its applications. The research in the field has also been predominantly motivated by the wide scale access to good quality and low cost 3D scanners, which has opened many new avenues. 
However, despite of all the recent advancements of algorithms for 3D meshes and point clouds, there is still room for enhancements to match the performance of the systems designed for 2D images on similar tasks. Nearly all the state-of-the-art methods in the various fields of computer vision are designed for 2D images and there are also immense differences in the sizes of training datasets for 2D images compared to any of their 3D counterparts. This points to a need for the development of bridging techniques, which can make efficient uses of both 2D images and 3D point clouds or meshes.  A technique that can fuse information from both 2D and 3D domains can benefit from the matured 2D vision as well as the contemporary advances in 3D vision. The complementary nature of 2D and 3D data can lead to an improved performance and efficiency in many potential applications, e.g., face recognition and verification, identification of objects or different regions of interest from coloured images in 3D maps, use of the 3D model of an object to locate it in an image in the presence of severe perspective distortion \citep{laga20183d}, as well as localization of a 2D images in a 3D point cloud. 

To move towards the goal of multi-domain information fusion from 2D and 3D domains, this paper proposes a novel approach to learn a framework to directly match feature descriptors extracted from 3D point clouds with those extracted from 2D images. This concept is used to localize images in point clouds directly generated from 3D scanners. To localize images, the developed framework is used to match 3D points of the point clouds to their corresponding pixels in 2D images with the help of multi-domain descriptor matching. The matched points between the images and the point clouds are then used to estimate the six degrees-of-freedom (DOF) position and orientation (pose) of images in the 3D world.

6-DOF pose estimation is an important research topic due to its numerous applications such as augmented reality, place recognition, robotic grasping, navigation, robotic pose estimation as well as simultaneous localization and mapping (SLAM). Current techniques for camera pose estimation can be classified into two major categories: \textbf{(i)} Regression networks based methods and \textbf{(ii)} Features based methods.  

The regression networks based methods (e.g. \cite{posenet,kendall2017geometric,walch2017image}), use deep neural networks to estimate the pose of the camera and consequently have high requirements for computational resources such as powerful GPUs and require a lot of training data \citep{xin2019review} from different viewpoints to ensure that the poses of query cameras are sufficiently close to the training ones \citep{sattler2019understanding}. Moreover they scale poorly with the increase in the size of 3D models and usually run into the problems of non-convergence for end-to-end training \citep{brachmann2018learning}. 

Features based method use hand-crafted approaches or deep learning methods to extract local or global features from images. An essential step in the state-of-the-art techniques in this category is the use of the Structure from Motion (SfM) pipeline \citep{colmapsfm,han2019image} to create sparse 3D models from images (e.g. \citep{li2010location,sattler2015hyperpoints,sattler2017efficient}). Structure from Motion pipelines provide a one-to-one correspondence between the points in the generated sparse point cloud and the pixels of the 2D images that were used to create the model. Several works use this information to localize images with respect to the 3D point cloud generated by SfM. However, model creation with SfM is a computationally expensive process which may be very time consuming depending on the number of images used and the quality of the point cloud to be generated. Models generated with SfM have especially poor quality for or miss out entirely on texture-less regions. Moreover, dependency on SfM generated point clouds renders such techniques futile for scenarios where point clouds have been obtained from 3D scanners. Now-a-days, high quality and user friendly 3D scanners (e.g. LIDAR, Microsoft Kinect, Matterport scanners and Faro 3D scanners) are available which can effectively render dense point clouds of large areas without the use of SfM. These point clouds are of better quality, not only because of their higher point density but also due to the reason that they can effectively capture bland surfaces that SfM based techniques tend to miss.

\begin{figure*}[t]
\begin{center}
   \includegraphics[width=1\linewidth]{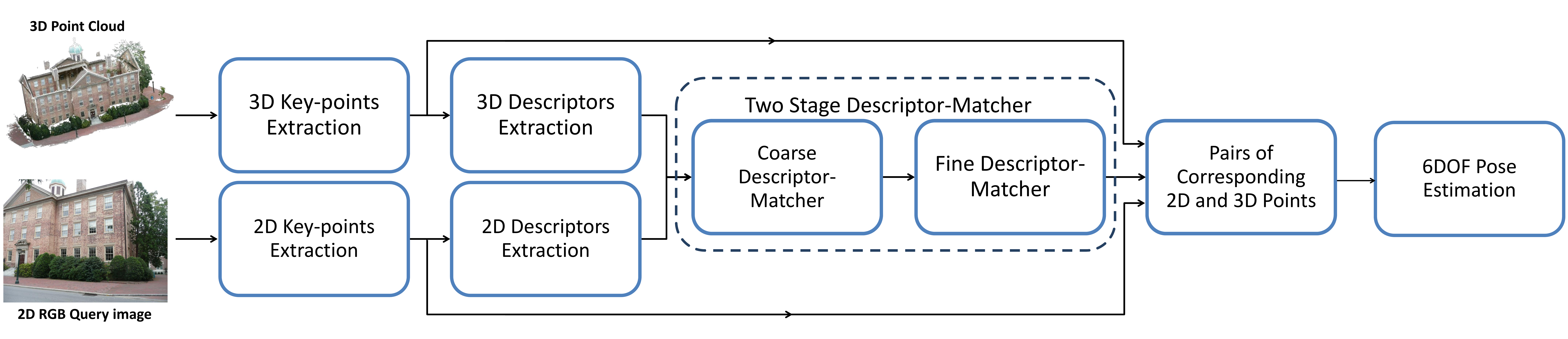}
\end{center}
   \caption{A block diagram of the test pipeline of the proposed technique. We extract 3D key-points and descriptors from the dense 3D Point Cloud. 2D key-points and their corresponding descriptors are extracted from the 2D RGB Query Image. Then our proposed `Descriptor-Matcher' algorithm directly matches the 2D descriptors with the 3D descriptors to generate correspondence between points in 2D image and 3D point cloud. This is then used with a robust pose estimation algorithm to estimate the 6-DOF pose of the query image in the 3D point cloud.}
\label{fig:test_chart}

\end{figure*}

To be able to directly localize 2D images in the point clouds generated from any 3D scanners, we propose a novel concept to directly match feature descriptors extracted from 3D point clouds with descriptors extracted from 2D images. The matched descriptors can then be used for 6-DOF camera localization of the image in the 3D point cloud. 
Figure \ref{fig:result} shows a section of the dense point cloud of the Shop Facade dataset \citep{posenet} along with a query image and the localization results of the proposed technique.
To match the feature descriptors from 2D and 3D domain, we generate a dataset of corresponding 3D and 2D descriptors to train a two-stage classifier called `Descriptor-Matcher'. For localization of a 2D image in a point cloud, we first use image feature extraction techniques such as Scale Invariant Feature Transform (SIFT) \citep{lowesift} to extract keypoints and their descriptors from the 2D image. Similarly, we use techniques designed for point clouds \citep{guo20143d} such as 3D-SIFT key-points \citep{lowesift,pcl}, 3D-Harris key-points \citep{harris,laga20183d} and Rotation Invariant Features Transform (RIFT) descriptors \citep{rift} to extract 3D keypoints and descriptors from the point cloud.
The Descriptor-Matcher is then used to find the matching pairs of 3D and 2D descriptors and their corresponding keypoints. This results in a list of coordinates in the point cloud and their corresponding pixels in the query image. The resulting one-to-one matches are then used in a robust algorithm for 6-DOF pose estimation to find the location and orientation of the query camera in the 3D point cloud. Figure \ref{fig:test_chart} shows the steps involved in our technique to estimate the camera pose for a query image.

A preliminary version of this work appeared in \cite{my2d3d}. To the best of our knowledge, \cite{my2d3d} was the first work: \textbf{(i)} to match directly 3D descriptors extracted from dense point clouds with the 2D descriptors from RGB images, and \textbf{(ii)} to use direct matching of 2D and 3D descriptors to localize camera pose with 6-DOF in dense 3D point clouds. This work extends \cite{my2d3d} by improving all the elements of the proposed technique, including dataset generation, Descriptor-Matcher and pose estimation. Additionally, we propose a reliable method to create a large collection of 2D and 3D points with matching locations in images and point cloud, respectively. We also present more details of the proposed method and also evaluate the proposed technique on more challenging datasets compared to \cite{my2d3d}.

The rest of this paper is organized as follows. Section \ref{related_work} discusses the various categories of the techniques in the literature for camera localization or geo-registration of images. The details of the proposed technique are presented in Section \ref{technique}. Section \ref{experiments} reports the experimental setup and a detailed evaluation of our results.
Finally, the paper is concluded in Section \ref{conclusion}.

\section{Related Work}\label{related_work}

The numerous applications of camera pose estimation and image localization render it as an interesting and active field of research. Traditionally, there are two distinct approaches to estimate the position and orientation of a camera \citep{xin2019review}: \textbf{(i)} Features based methods and \textbf{(ii)} Network based pose regression methods. The proposed method forms a new category of possible approaches: \textbf{(iii)} Direct 2D-3D descriptor matching based methods.

\subsection{Features based methods}  Features based methods extract convolutional or hand-crafted features from 2D images and use them in different manners to localize the images. However, many of these methods only estimate approximate locations and use number of inliers found with RANSAC \citep{ransac} as a criterion for image registration e.g., a query image is considered as registered if the RANSAC stage of the method can find 12 inliers among the features for the query image. This is partly due to the fact that some datasets for image localization do not provide the ground truth position and orientation information for the query cameras. However, inlier count is not a reliable criterion and it does not represent the actual performance of any given method.  Feature based methods can further be classified into two types of methods: Image retrieval based methods and SfM based methods.

\subsubsection{Image retrieval based methods } 
Image retrieval based methods involve the use of a large geo-tagged database of images. To localize a 2D query image, these methods use different types of features to retrieve images similar to the query image. The average of the retrieved database images can be used as the predicted location of the query image. Alternatively, the poses of the retrieved images can be used to triangulate the pose of the query camera \citep{chen2011city,zamir2010accurate}. These methods cannot be used for the localization of 2D images in point clouds. Also, many times the images in the dataset are not sufficiently close to the query image which results in significant errors in pose estimation.

\subsubsection{SfM-based methods} 
SfM-based methods produce better pose estimates than image retrieval based methods \citep{sattler2017efficient}. These methods use the SfM pipeline \citep{colmapsfm}. SfM first extracts and matches features such as SIFT, SURF or ORB, from the set of training images. Then the matched features between the different 2D images are used to create a 3D model on an arbitrary scale. Each point in the 3D model is created by triangulation of points from multiple images.

\cite{irschara2009structure} used image retrieval to extract 2D images that were similar to the query 2D image from the training database. The extracted images were used with the SfM model to improve the accuracy of the estimated camera poses. 
\cite{li2010location} compared the features extracted from the query 2D images with the 2D features corresponding to the points in the SfM model to localize the images.
\cite{sattler2015hyperpoints} improved the localization process with a visual vocabulary of 16 million words created from the features of the database images and their locations in the SfM point cloud.
Later, \cite{sattler2017efficient} further extended their work with the help of a prioritized matching system to improve the estimated poses for the query images.  

However, these methods can only work with point clouds generated with SfM based pipelines \citep{piasco2018survey}. This is mainly due to the reason that the sparse point clouds generated with SfM save the corresponding information of the points and features from 2D images that were used to create the sparse SfM point cloud. SfM based methods rely on this information for pose estimation. Also some works use element-wise mean (or any other suitable function) of the feature descriptors of the points that were used to create a 3D point in the SfM point cloud as a 3D descriptor of that 3D point. Such an approximation of 3D features is dependent on the inherent information in the point cloud generated with SfM, which is not available if the point cloud had to be generated from a 3D scanner. 

Structure from Motion is a computationally expensive and time consuming process. The point clouds generated with SfM are sparse and very noisy \citep{feng20192d3d}. SfM models have especially poor quality at bland or texture-less regions and may miss out such areas altogether. Moreover, it requires multiple images of the same area from many different angles for good results, which is practically a difficult requirement, especially for large areas or buildings. Also the generated models are created on an arbitrary scale and it is not possible to determine the exact size of the model without extra information from other sources \citep{feng20192d3d}. The availability of high quality 3D scanners has made it possible to capture large scale point clouds in an efficient manner without the need to capture thousands of images to be used for the SfM pipeline. Moreover, LIDAR and other 3D scanners are now becoming an essential part of robots, particularly for locomotion and grasping. Therefore, it is essential to develop methods that can estimate 6-DOF poses for cameras in point clouds captured from any scanner.

\subsection{Network-based pose regression methods} 
These methods use deep neural networks to estimate the position and orientation of query images through pose regression. However, \cite{sattler2019understanding} showed that these methods can only produce good results when the poses of the query images are sufficiently close to the training images.
\cite{posenet} proposed a convolutional neural network, called PoseNet, which was trained to regress the 6-DOF camera pose of the query image. 
\cite{kendall2017geometric} later improved the loss function in Posenet while \cite{walch2017image} improved the network architecture to reduce the errors in pose estimations. 
\cite{brachmann2017dsac} introduced the concept of differentiable RANSAC. Instead of direct pose regression through a neural network, they created a differential version of RANSAC \citep{ransac} to estimate the location and orientation of the query cameras. Specifically, model of \cite{brachmann2017dsac} was composed of two CNNs, one to predict scene coordinates and the other CNN to select a camera pose from a pool of hypotheses generated from the output of the first CNN. However, their method was prone to over-fitting and did not always converge during end-to-end-optimization especially for outdoor scenes \citep{brachmann2018learning}. 
\cite{radwan2018vlocnet++} used a deep learning based architecture to simultaneously carry out camera pose estimation, semantic segmentation and odometry estimation by exploiting the inter-dependencies of these tasks for assisting each other. 
 
\cite{brachmann2018learning} modified \citep{brachmann2017dsac} with a fully convolutional network for scene coordinate regression as the only learnable component in their system. Soft inlier count was used to test the pose hypotheses. Although it improved the localization accuracy, the system still failed to produce results for datasets with large scale.

\subsection{Direct 2D-3D descriptor matching based methods}
Our preliminary work \citep{my2d3d} was the first technique to estimate the 6-DOF pose for query cameras by directly matching features extracted from 2D images and 3D point clouds. \cite{feng20192d3d} trained a deep convolutional network with triplet loss to estimate descriptors for patches extracted from images and point clouds. The estimated patches were used in an exhaustive feature matching algorithm for pose estimation. However, their system achieved a limited localization capability. 

This paper improves the concept of \citep{my2d3d} with an unconstrained method to extract pairs of corresponding 2D and 3D points for training, and improves the structure of Descriptor-Matcher. It also introduces a better pose estimation strategy. Our technique directly extracts 3D key-points and feature descriptors from point clouds which, in contrast to SfM based approaches, enables us to estimate poses in point clouds generated from any 3D scanner. Moreover, the capability of our method to work with dense point clouds allows it to use better quality feature descriptors, which additionally helps in the pose estimation of the query images.

\section{Proposed Technique}\label{technique}
The direct matching of 2D and 3D descriptors in our technique provides a way to localize the position and orientation of 2D query images in point clouds which is not constrained by the method of point cloud generation, the type of 3D scanners used or the indoor and outdoor nature of the environment. 

At the core of our technique is a classifier, called `Descriptor-Matcher', which is used to match the feature descriptors from 2D and 3D domain. Descriptor-Matcher is composed of two stages: coarse and fine. To train the Descriptor-Matcher, we need a dataset of 3D points and their corresponding pixel locations in the training images. It is impractical to manually create a dataset of matching 2D and 3D points, especially for large scales. To overcome this issue, we propose a method to automatically collect a large number of matching 3D and 2D points and their corresponding descriptors from a point cloud and a given set of 2D training images. The trained Descriptor-Matcher can then be used to localize a 2D query image in the point cloud.

\subsection{Extraction of Key-points and Feature Descriptors}
\label{technique:keypoint_extraction}

Let us assume a dataset with $N$ number of training images with known ground truth poses. 
Let $PC_d$ be the point cloud in which the query images need to be localized: 
\begin{equation}
\label{e1}
PC_{d}= \bigcup_{i=1}^{Np}(x_i,y_i,z_i)
\end{equation} 
where $\bigcup$ is the union operator, $Np$ is the number of points in the point cloud and $(x_i,y_i,z_i)$ are the Cartesian coordinates of the $i^{th}$ point of the point cloud. Nowadays, most of the 3D sensors can produce RGB images with ground truth camera positions relative to the generated point clouds. In case such information is not available, Multi View Stereo \citep{colmapmvs} can be used to generate ground truth camera poses for the training images as explained in \citep{my2d3d}.

To create the dataset for training, we first use methods designed for point clouds to extract 3D keypoints and feature descriptors from the point cloud $PC_d$ \citep{pcl,guo20143d}. This results in a set of 3D keypoints: 
\begin{equation}
\label{e2}
keys3D= \bigcup_{j=1}^{N3}(x_j,y_j,z_j)
\end{equation}
and their corresponding 3D descriptors:
\begin{equation}
\label{e3}
desc3D= \bigcup_{j=1}^{N3}(u_j^1,u_j^2,u_j^3,...u_j^{p})
\end{equation}
where $(u_j^1,u_j^2,u_j^3,...u_j^{p})$ is the $p$ dimensional 3D descriptor of the $j^{th}$ 3D keypoint  with Cartesian coordinates $(x_j,y_j,z_j)$, and $N3$ is the number of the detected keypoints in the dense point cloud.

Similarly, we use keypoint and descriptor extraction methods specific for images to get a set of 2D keypoints $keys2d$ and feature descriptors $desc2d$ for each training image: 
\begin{equation}
\label{e4}
keys2D_n= \bigcup_{k=1}^{N2_n}(\widetilde{x}_{k,n},\widetilde{y}_{k,n})
\end{equation}
\begin{equation}
\label{e5}
desc2D_n= \bigcup_{k=1}^{N2_n}(v_{k,n}^1,v_{k,n}^2,v_{k,n}^3,...v_{k,n}^{q})
\end{equation}
where $N2_n$ is the number of the detected keypoints in $n^{th}$ image. $\widetilde{x}_{k,n}$ and $\widetilde{y}_{k,n}$ are the horizontal and vertical pixel coordinates, respectively,  of the $k^{th}$ keypoint in the $n^{th}$ training image and $(v_{k,n}^1,v_{k,n}^2,v_{k,n}^3,...v_{k,n}^{q})$ is the $q$ dimensional 2D descriptor of that keypoint.

\subsection{Dataset Generation}
\label{technique:dataset}

\subsubsection{Back ray Tracing}
To calculate the distance between 2D and 3D keypoints, we need to either project the 3D keypoints onto the 2D image or the 2D keypoints onto the 3D point cloud. 
Projecting the 2D keypoints will involve the voxelization of the point cloud which is a computationally expensive process compared to its alternative (i.e., back ray tracing) for the task at hand. Therefore, for each image in the training set, we use the intrinsic and extrinsic matrix of the camera to trace back rays from the 3D key points $keys_{3d}$ on the image. Mathematically, back ray tracing can be represented through the following equation:
\begin{equation}
\label{e6}
\begin{aligned}
\begin{bmatrix}  \hat{s}_{j,n}\, \hat{x}_{j,n},\,\hat{s}_{j,n}\, \hat{y}_{j,n},\,\hat{s}_{j,n} \end{bmatrix}^{tr} = \\  \begin{bmatrix}K_n\end{bmatrix}  \!  \begin{bmatrix}R_n|T_n\end{bmatrix}  \! \begin{bmatrix}x_j,y_j,z_j,1\end{bmatrix}^{tr}
  \\  \forall j=1,2,3,...,N3
\end{aligned}
\end{equation}
Where $K_n$ is the intrinsic matrix of the camera, $R_n$ is the rotation matrix and $T_n$ is the translation vector for the conversion of points from world coordinates to camera coordinates for the $n^{th}$ image. The $tr$ superscript denotes the transpose of a matrix. The $\hat{x}_{j,n}$ and $\hat{y}_{j,n}$ are the horizontal and vertical pixel coordinates, respectively, of the projected 3D keypoint on the $n^{th}$ image and $\hat{s}_{j,n}$ is the depth of the 3D keypoint from the camera. We keep only those projected keypoints that are within the boundaries of the image and have a positive depth value so that they are in front of the camera:
 \begin{equation}\label{e7}
  \begin{aligned}
\Theta_n=  \bigcup(\hat{x}_{j,n}, & \hat{y}_{j,n},\hat{s}_{j,n}) \   s.t. \  (0<\hat{x}_{j,n}<img\_w_n)\  \\ and & \  (0<\hat{y}_{j,n}<img\_h_n)\ and\ (\hat{s}_{j,n}>0) \\ & \quad \qquad \qquad \qquad \forall \  j=1,2,3,...,N3
 \end{aligned}
\end{equation}
Where $img\_w_n$ and $img\_h_n$ are the width and height of the $n^{th}$ image, respectively. However, as keypoints are extremely sparse, it is possible that points from the areas of the point cloud that are not visible in the image, get projected on the image. For example, if the camera is looking at a wall at a right angle and there is another wall behind the first one, then it is possible that the keypoints detected on the back wall get projected on the image. Theses false projections can create problems during the training of the Descriptor-Matcher as they result in the matching of locations between images and point clouds which are not the same and increase the number of false positives. Figure \ref{fig:depth_block_filter} shows examples of false projections of 3D keypoints from locations in point cloud which are not visible in the 2D image.

\subsubsection{Depth Block Filtering}
\label{technique:dataset:DBF}
To overcome this problem, we devised a strategy based on the 3D points of the point cloud $PC_d$, called Depth Block Filtering.
Similar to the 3D keypoints, we back trace the points in $PC_d$ to the image with the help of the intrinsic matrix of the camera, $K_n$, and the augmented matrix created by appending the rotation matrix $R_n$ with the translation vector $T_n$ to get the pixel coordinates of the 3D points:
\begin{equation}
\label{e8}
\begin{aligned}
\begin{bmatrix} \hat{s}_{i,n}\,\hat{x}_{i,n},\,\hat{s}_{i,n}\,\hat{y}_{i,n},\,\hat{s}_{i,n} \end{bmatrix} ^{tr}\!=\\ \begin{bmatrix}K_n\end{bmatrix}    \! \begin{bmatrix}R_n|T_n\end{bmatrix}  \! \begin{bmatrix}x_i,y_i,z_i,1\end{bmatrix}^{tr}
  \\  \forall i=1,2,3,...,Np
\end{aligned}
\end{equation}

where $\hat{x}_{i,n}$ and $\hat{y}_{i,n}$ are the horizontal and vertical pixel coordinates, respectively, of the projected points of the 3D point cloud on the $n^{th}$ image and $\hat{s}_{i,n}$ is the depth of the projected point from the camera.
We filter out all those points that are outside the boundaries of the image and behind the camera i.e., those with a negative depth value $\hat{s}_{i,n}$.
 \begin{equation}\label{e9}
  \begin{aligned}
\Gamma_n=  \bigcup(\hat{x}_{i,n}, & \hat{y}_{i,n},\hat{s}_{i,n}) \   s.t. \  (0<\hat{x}_{i,n}<img\_w_n)\  \\ and & \  (0<\hat{y}_{i,n}<img\_h_n)\  and\ (\hat{s}_{i,n}>0) \\ & \quad \qquad \qquad\qquad \forall \  i=1,2,3,...,Np
 \end{aligned}
\end{equation}

\begin{figure*}[htbp]
\begin{subfigure}{.5\textwidth}
  \centering
  \includegraphics[width=1\linewidth]{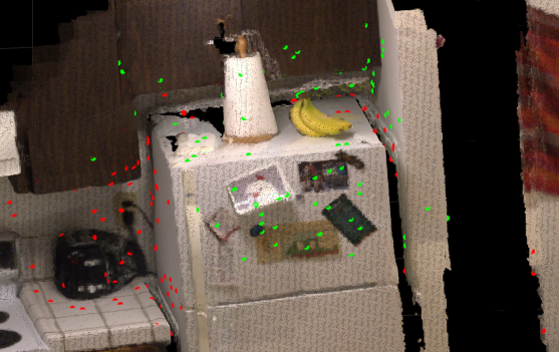}  
  \captionsetup{justification=centering}
 
  \caption{}
  \label{fig:3a}
\end{subfigure}
\begin{subfigure}{.5\textwidth}
  \centering
  \includegraphics[width=1\linewidth]{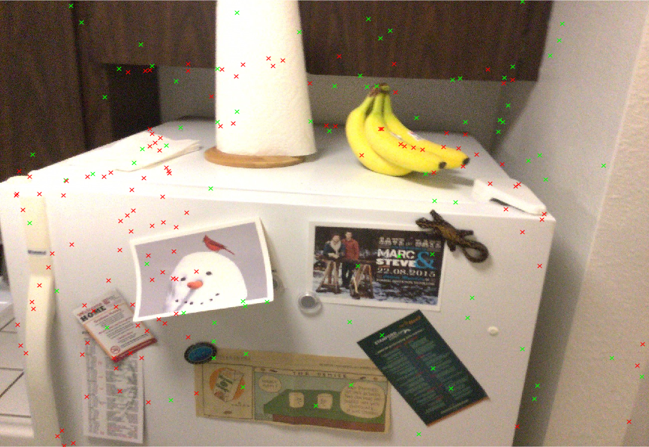}  
  \captionsetup{justification=centering}
  \caption{}
  \label{fig:3b}
\end{subfigure}

\begin{subfigure}{.5\textwidth}
  \centering
  \includegraphics[width=1\linewidth]{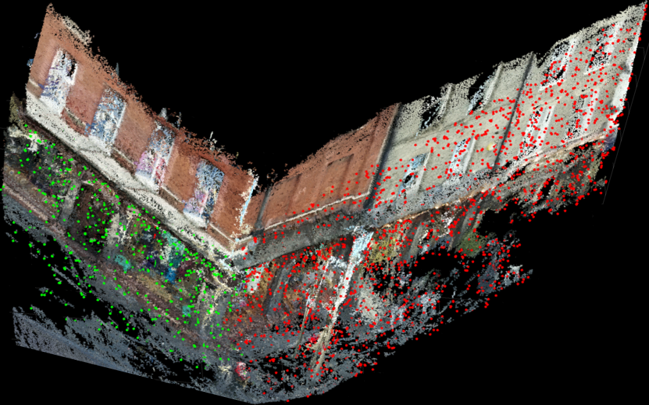}  
  \captionsetup{justification=centering}
  \caption{}
  \label{fig:3c}
\end{subfigure}
\begin{subfigure}{.5\textwidth}
  \centering
  \includegraphics[width=1\linewidth]{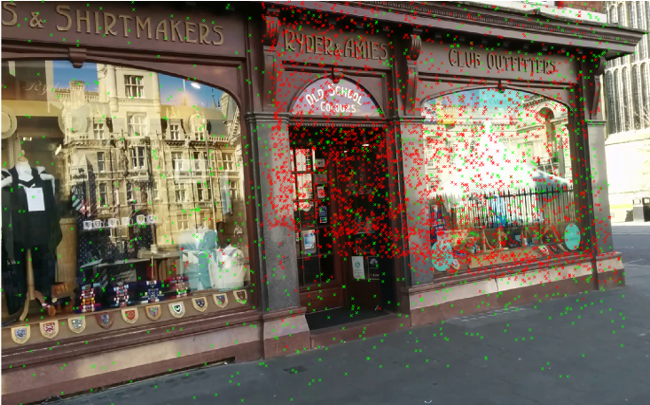}  
  \captionsetup{justification=centering}
  \caption{}
  \label{fig:3d}
\end{subfigure}

\caption{Examples of results from Depth Block Filtering (DBF). Due to the sparse nature of 3D keypoints, even the points occluded from the camera get projected by back ray tracing e.g., keypoints occluded by the fridge in (a) and (b) and points from around the shop in (c) and (d). Depth Block Filtering effectively removes the occluded keypoints and only retains the keypoints in the direct line of sight of the camera. \textbf{Left column:} Snapshot of sections from point clouds with 3D keypoints (red + green) that get projected on a training image. \textbf{Right column:} Image with the projected locations of 3D keypoints. \textbf{Red points:} Keypoints removed by DBF as occluded points. \textbf{Green points:} Keypoints retained by DBF. See Section \ref{technique:dataset:DBF} for details of DBF.}
\label{fig:depth_block_filter}
\end{figure*}

We convert the projected pixel values $\hat{x_i}$ and $\hat{y_i}$ in $\Gamma_n$ to whole numbers. Any duplicates in the resulting set that have the same values for both $\hat{x_i}$ and $\hat{y_i}$ are removed by keeping only those elements of $\Gamma_n$ which are closest to the position of the camera i.e., among the duplicate pixel locations, we keep the triple with the minimum value of $\hat{s_i}$. The depth values $\hat{s_i}$ in $\Gamma_n$ are then used to create a depth map for the training image. 
Next, we use the generated depth map to create blocks of size $\tau \times \tau$ pixels around the locations of the triples in the set of projected 3D keypoints $\Theta_n$.
Finally, we filter out all the triples in $\Theta_n$ whose depth values $\hat{s_j}$ are greater than a threshold $\varphi$ in their respective depth block, where $\varphi$ and $\tau$ are constants. This strategy helps to cater not only for wrong projections but also for any holes in the point cloud as well. Figure \ref{fig:depth_block_filter} shows examples of the results from depth block filtering.

\subsubsection{Creating 2D-3D Correspondences}
We then calculate the Euclidean distances between the remaining projected points $(\hat{x}_{j,n},\hat{y}_{j,n})$ in the set of projections of 3D keypoints $\Theta_n$ and the 2D keypoints $keys2D_n$ on the image. The pairs of 2D and 3D keypoints whose projections are within a specified distance $\alpha$, measured in the units of pixels, are considered as the matching pairs. In the case that there are more than one 2D keypoints within the specified distance of the projection of a 3D keypoint or multiple 3D keypoints' projections are close to a 2D keypoint, only the pair with the smallest distance is considered for the dataset, i.e., we treat the matching of keypoints as a one-to-one correspondence for each image.
Let $\zeta_n$ be the set of the pair of indices, corresponding to the points of $N3$ and $N2_n$, that are within an error threshold $\alpha$:
 \begin{equation}\label{e10}
  \begin{aligned}
\zeta_n=  \bigcup(j,k) \quad \forall  \left || (\hat{x_j}-\widetilde{x}_{k,n}),(\hat{y_j}-\widetilde{y}_{k,n}) \right || < \alpha \\ \quad   where \quad j=1,2,3,...,N3, \\  k=1,2,3,...,N2 
 \end{aligned}
\end{equation}

We use $\zeta_n$ to create a matching set of 3D and 2D points by retrieving the keypoints from $keys3D$ and $keys2D_n$ according to the indexes. This process is repeated for each training image and then the resulting sets of matching keypoints are concatenated to create a dataset of matching 3D and 2D points for the whole training set. 

To obtain a dataset of corresponding 3D and 2D descriptors for the matching points, we retrieve the corresponding descriptors for the 3D key-points from $desc3D$ and the 2D descriptors from $desc2D$. 
It is to be noted that in the resulting dataset, one 3D descriptor can correspond to multiple 2D descriptors as one 3D keypoint can appear in multiple images of the same place taken from different poses.

\subsection{Training}
\label{technique:train}
The generated dataset of corresponding 2D and 3D features is used to train the Descriptor-Matcher. The Descriptor-Matcher is composed of two stages: a coarse matcher and a fine  matcher, connected in series (see Figure \ref{fig:test_chart}). The two stages allow the Descriptor-Matcher to maintain both a high precision and computational efficiency at the same time, as explained in Section \ref{technique:test}.
We evaluated several different classifiers for the coarse and fine stages including multi-layer fully connected Neural Networks \cite{khan2018guide}, Support Vector Machines with different kernels, Nearest Neighbour Classifiers with dictionary learning and Discriminant Analysis. However, we empirically found that Classification Tree \citep{treeref} performs best for the coarse stage of the matcher, while for the fine stage only Random Forest classifier \citep{randomforest} produces suitable results. Moreover, these classifiers have better robustness, speed, generalization capability and ability to handle over-fitting for the task at hand.

We treat the problem of matching the descriptors from images and point cloud as a binary classification problem. At the input of the Descriptor-Matcher, we provide 2D and 3D descriptors, concatenated in the form of a $(p\! +\! q)\! \times \! 1$ vector. A positive result at the output indicates that the concatenated descriptors are from the same location in the image and the point cloud, while a negative result shows a mismatch. 
To split nodes in the classification tree of coarse matcher, as well as in the trees in Random Forest, we used Gini's Diversity Index to measure the impurity of nodes. The impurity criterion for our case is defined as:
\begin{equation}\label{eq:gdi_binary}
Gini's\ index=\frac{2\times r^+ \times r^-}{(r^++r^-)^2}
\end{equation}
where $r^+$ and $r^-$ are the number of positive and negative samples, respectively, at any given node of a tree. Consequently, an impure node will have a positive value for the splitting criterion, while a node with only positive or only negative samples will have Gini's diversity index equal to zero.

To train the classifiers, the corresponding descriptors of the matching 3D and 2D keypoints in the generated dataset (Section \ref{technique:dataset}) were concatenated to create positive training samples. For the negative samples, we concatenated the 2D and 3D descriptors of the non-matching points in the dataset. We define non-matching points as those pairs whose corresponding locations in the point cloud are 
at least a distance $\beta$ apart. However, with the increase in the scale of the point cloud, it is possible that there are regions with similar appearance at different locations in the point cloud. This is particularly a concern for indoor scenarios, where there are comparatively more bald regions and repeated structures. To overcome this problem we also ensured that the Euclidean distance between the descriptors of non-matching points is greater than a threshold $\gamma$. We only used a randomly selected subset of the generated negative samples for training due to the large number of possible one-to-one correspondences between non-matching pairs. We optimized the fine-matcher for maximum precision to minimize the number of false positives in the final matches.

\subsection{Testing}
\label{technique:test}
At test time, we first extract key-points and descriptors from the 2D query image that needs to be localized in the point cloud. Similarly, 3D keypoints and descriptors are extracted from the point cloud, if not already available. Then we concatenate the 2D descriptors of the query image and the 3D descriptors of the point cloud in a one-to-one fashion and use the two stage Descriptor-Matcher to find the matching and non-matching pairs. The pairs of descriptors are first tested by the coarse matcher and only the pairs with positive results for matches are passed to the fine-matcher. As the coarse matcher is composed of a single classification tree, it greatly reduces the number of pairs which need to be tested by the fine matcher, at only a small computational cost, thus greatly decreasing the prediction time of the algorithm.  
Then, we retrieve the corresponding keypoints for the descriptor pairs positively matched by the fine-matcher and use them as matching locations from the image and the point cloud for pose estimation.

However, just like any classifier, the output of the Descriptor-Matcher contains some false positives and the matched 3D and 2D points are not completely free of wrong matches. We use two conditions to improve the descriptor matching.
\textbf{First}, we use the probability scores for a positive match (which are in the range $ 0-1$) as the confidence value that a 2D descriptor and a 3D descriptor represent the same locations in the quey image and the point cloud, respectively.
For a positive match, the predicted confidence value must be greater than $0.5$. \textbf{Second}, we also use two-way matching to improve the reliability of the matches. Specifically, based on the highest prediction confidence, we find the closest matching 3D descriptor for each 2D descriptor, and the closest matching 2D descriptor for each 3D descriptor. For a 2D descriptor to be treated as a corresponding pair for a 3D descriptor, the confidence value of the predicted match must be greater than the confidence value of that specific 2D descriptor matching with any other 3D descriptor and vice versa.  

To further filter out the false positives we use the MLESAC \citep{mlesacmatlab} algorithm along with the P3P \citep{p3pmatlab} algorithm to find the best set of points for the estimation of the position and orientation of the camera for the query image. 
MLESAC is an improved and more generalized algorithm compared to the traditional RANSAC \citep{ransac}. MLESAC generates the tentative solutions in the same manner as RANSAC. However, it produces a better final solution as, in addition to maximizing the number of inliers in the solution, it uses maximum likelihood estimation based on a Gaussian distribution of noise to minimize the re-projection error between the 2D image and 3D points \citep{mlesacmatlab}. 
Figure \ref{fig:test_chart} shows a block diagram of the steps involved to localize a 2D query image in a 3D point cloud. The predicted pose of the camera can further be refined with the application of the R1PPnP pose estimation algorithm \citep{zhou2018re} on the points identified as inliers by the MLESAC algorithm to estimate the final position and viewing direction of the query camera in the point cloud. The predicted pose information can be used to extract the section of the point cloud that corresponds to the query image for visualization purposes (Figure \ref{fig:result}).

\section{Experiments and Analysis}\label{experiments}
To evaluate the performance of our technique, we carried out extensive experiments on a number of publicly available datasets. These include Shop Facade \citep{posenet}, Old Hospital \citep{posenet}, Trinity Great Court \citep{kendall2017geometric}, King's College \citep{posenet} and St. Mary Church \citep{posenet} 
datasets from the Cambridge Landmarks Database for \textbf{outdoor scenarios}. 
To test the localization capability of the proposed technique for \textbf{indoor cases}, we used the Kitchen \citep{valentin2016learning} and Living Room \citep{valentin2016learning} Datasets from the Stanford RGB Localization Database and Baidu Indoor Localization Dataset \citep{idl}.

In our experiments, SIFT key-points and descriptors \citep{lowesift} were extracted from the 2D query images. For the point clouds, we used 3D SIFT key-points \citep{pcl} and 3D RIFT descriptors \citep{rift} to extract keypoints and feature descriptors, respectively. 3D SIFT is a key-point extraction method designed for point clouds, which is inspired from the 2D SIFT key-point extraction algorithm \citep{lowesift} for 2D images. The original 2D SIFT algorithm was adapted for 3D point clouds by substituting the function of intensity of pixels in an image with the principal curvature of points in a point cloud \citep{pcl}.

We set the maximum distance between the image keypoints and the projected 3D keypoints $\alpha=5\ pixels$ for the generation of positive samples. For the formation of one-to-one pairs for negative samples, we set the minimum distance between the 3D keypoints, $\beta$, to 0.5 and the minimum Euclidean distance between the 3D descriptors, $\gamma$ to $0.3$.
To optimize the parameters for coarse and fine matchers, we divided the generated dataset of corresponding 2D and 3D descriptors (see Section \ref{technique:dataset}) into training and validation sets in the ratio of $8\! :\! 2$. Based on the performance on the validation set, we used grid search \citep{gridsearch} to fine tune the classification cost and the maximum number of splits in a tree for both coarse and fine matchers, as well as the number of trees in the Random Forest. 
Finally, the complete dataset of matching 2D and 3D features was used to train the Descriptor-Matcher with the optimized parameters.

\subsection{Evaluation Metrics}
We use the positional and the rotational errors in the predicted poses of the query images to evaluate the performance of our technique. As such, the positional error is calculated as the Euclidean distance between the ground truth and the predicted positions of the query camera in the 3D point cloud:
\begin{equation}
    \label{pos_err}
    position\_error\! =\!  \left || (x_g-x_e),(y_g-y_e),(z_g-z_e) \right |\! |
\end{equation}
where $(x_g,y_g,z_g)$ and $(x_e,y_e,z_e)$ are the ground truth and estimated positions of the query image's camera in the 3D point cloud, respectively. 

For the estimation of the rotational error, we calculated the minimum angle between the viewing directions of the predicted and the ground truth cameras for the 2D query image. If the predicted rotation matrix is represented by $R_e$ and the ground truth rotation matrix is $R_{g}$, then the rotational error $\phi$ in degrees can be calculated as follows:
\begin{equation}
\label{eq:R_err}
    \phi = \frac{\pi}{180} \times cos^{-1}(\frac{trace(R_g \times R_e^{tr})-1}{2})
\end{equation}

\begin{sidewaystable*}[htbp]%
\caption{Median and percentile localization errors for the outdoor datasets: Shop Facade, Old Hospital, St. Mary Church, King's College and Great Court Datasets. P stands for percentile, e.g., P 25\% means the maximum error for 25\% of the data when errors are sorted in the ascending order. m stands for metres}
  \centering
  
    \begin{tabular}{|l|l|c|c|c|c|c|}
    \specialrule{.15em}{.0em}{.0em} 
    \textbf{Outdoor Datasets $\downarrow$} & \textbf{Errors $\downarrow$ \textbackslash{} Metrics $\rightarrow$} & \textbf{Median} & \textbf{P 25\%} & \textbf{P 50\%} & \textbf{P 75\%} & \textbf{P 90\%} \\\specialrule{.15em}{.0em}{.0em} 
    
    \multirow{2}[0]{*}{\textbf{Shop Facade}} & \textbf{Position Error (m)} & 0.0860 m & 0.0307 m & 0.0860 m & 0.4258 m & 3.3890 m \\\cline{2-7}
          & \textbf{Angle Error (degrees)} & 0.7792\degree & 0.2776\degree & 0.7792\degree & 4.2890\degree & 33.6595\degree \\\specialrule{.15em}{.0em}{.0em}
    
    \multirow{2}[0]{*}{\textbf{Old Hospital}} & \textbf{Position Error (m)} & 0.1295 m &  0.0649 m & 0.1295 m & 0.2561 m & 3.2744 m\\\cline{2-7}
          & \textbf{Angle Error (degrees)} & 0.2210\degree & 0.1312\degree & 0.2210\degree & 0.6153\degree & 5.6263\degree \\\specialrule{.15em}{.0em}{.0em}
    \multirow{2}[0]{*}{\textbf{St. Mary Church}} & \textbf{Position Error (m)} & 0.1479 m & 0.0689 m & 0.1479 m & 0.9518 m & 13.3455 m  \\\cline{2-7}
          & \textbf{Angle Error (degrees)} &0.4671\degree &  0.1814\degree & 0.4671\degree & 3.5267\degree & 35.3109\degree\\\specialrule{.15em}{.0em}{.0em}
          \multirow{2}[0]{*}{\textbf{King's College}} & \textbf{Position Error (m)} & 0.0877 m & 0.0533 m & 0.0877 m & 0.1332 m & 0.2108 m\\\cline{2-7}
          & \textbf{Angle Error (degrees)} & 0.1476\degree & 0.0880\degree & 0.1476\degree & 0.2613\degree & 0.4694\degree \\\specialrule{.15em}{.0em}{.0em}

    \multirow{2}[0]{*}{\textbf{Great Court}} & \textbf{Position Error (m)} & 0.5098 m & 0.2520 m & 0.5098 m & 1.5401 m & 14.7130
 m\\\cline{2-7}
          & \textbf{Angle Error (degrees)} & 0.3526\degree & 0.1489\degree & 0.3526\degree & 1.3117\degree & 12.8685\degree \\\specialrule{.15em}{.0em}{.0em}

    \end{tabular}
    \label{tab:shop}
    \label{tab:hospital}
    \label{tab:church}
       \label{tab:king}
    \label{tab:court}
    \label{tab:street}
\end{sidewaystable*}

\subsection{Outdoor Datasets}
We used the datasets from the Cambridge Landmarks Database \citep{posenet,kendall2017geometric} to test the localization performance of the proposed method in the outdoor scenarios.
The images in these datasets were captured at different times under different lighting and weather conditions and contain a lot of urban clutter which increases the challenges for precise localization of camera poses.
As the Cambridge Landmarks Database does not contain dense point clouds, we used the COLMAP's Multi View Stereo Pipeline \citep{colmapmvs} to generate dense point clouds for these datasets. 
Table \ref{tab:shop} reports the median errors for the estimated positions and orientations of the cameras for the query images for the outdoor datasets. We also report the percentile errors for the 25\%, 50\%, 75\% and 90\% of the query images in these datasets for the estimated camera poses.

\subsubsection{Shop Facade Dataset}
The Shop Facade Dataset \citep{posenet} from the Cambridge Landmarks Database is composed of the images of the intersection of two streets in Cambridge. The images mainly focus on the shops at the intersection. It covers an area of more than $900\ m^2$. It contains a total of 334 images with 103 images in the query set. We used the standard train-test split as defined by the authors of the dataset. Our proposed technique was able to localize all the images in the query set.

\subsubsection{Old Hospital Dataset}
The Old Hospital dataset \citep{posenet} contains 1077 images. There are 182 query images in the train-test split defined by the dataset's authors. The dataset covers an area of $2000\ m^2$. This dataset suffers particularly from the challenges of repetitive patterns, as well as high symmetry due to similar constructions on both sides of the centre of the building. The localization results are shown in Table \ref{tab:hospital}.

\subsubsection{St. Mary Church Dataset}
The St. Mary Church Dataset \citep{posenet} is composed of 2017 images of the Great St. Mary Church in Cambridge. It encompasses an area of $4800\ m^2$. Many of the images contain occlusions caused by pedestrians and other urban clutter due to which it becomes challenging to localize images in the 3D point cloud. We used the query set defined by the authors of the dataset which contains $530$ images for the evaluation of our technique, while the remaining 2D images were used to train the Descriptor-Matcher. Our technique successfully localized all the images in the query set.

\subsubsection{King's College Dataset}
This dataset covers the location and building of the King's College, which is one of the constituent colleges of the University of Cambridge \citep{posenet}. It consists of 1563 images captured with the camera of a smart phone. King's College covers an area of more than $5600\ m^2$. We used the train-query split of images defined by the authors of the dataset. There are 343 images in the query set. Table \ref{tab:king} shows the results of our technique on the dataset.

\subsubsection{Trinity Great Court Dataset}
Trinity Great Court is the main court (courtyard) of Trinity College, Cambridge. It is one of the largest enclosed courtyards in Europe with an area of $8000\ m^2$. The train and test sets consist of $1532$ and $760$ number of images, respectively \citep{kendall2017geometric}. The proposed technique successfully localized all the images in the dataset.

\subsection{Indoor Datasets}
Most of the pose estimation techniques in the literature have either been evaluated only for outdoor scenarios or for very small indoor scenes (e.g. tested for a maximum volume of $6m^3$ on Seven Scenes Database \citep{sevenscenes} ). 
Indoor localization of images is a more challenging problem compared to the outdoor settings \citep{idl}. Due to similar items (e.g., furniture) and the same construction patterns (e.g., cubicle shape of rooms, similar patterns on floor or ceiling, stairs), it is possible that different regions of the building look very similar, which greatly increases the possibility of wrong matches and incorrect camera pose estimation. On the other hand, indoor localization is a more useful application of image localization as  many of the traditional localization methods, such as GPS based localization, do not work properly in indoor settings. To evaluate the performance of our technique on practical indoor localization scenarios, we used the Kitchen and Living Room Datasets from the Stanford RGB Localization Database \citep{valentin2016learning} and Baidu Indoor Localization dataset \citep{idl}.
Table \ref{tab:idl} shows the median positional and rotational errors along with the percentile errors for the intervals of 25\%, 50\%, 75\%, and 90\% data on the indoor datasets.

\subsubsection{Kitchen Dataset}
This dataset contains RGB images and a 3D model of a Kitchen in an apartment. It is part of the Stanford RGB Localization Database \citep{valentin2016learning}. It covers a total volume of $33m^3$. The 2D images and the 3D point cloud were created with a Structure.io\footnote{https://structure.io/structure-sensor} 3D sensor coupled with an iPad. Both sensors were calibrated and temporally synced. We randomly selected $20\%$ of the images for the query set while used the remaining images for training the Descriptor-Matcher. Our technique successfully localized all the query images in the 3D model with high accuracy. We were able to estimate the 6-DOF poses of more than 90\% of the images with errors less than 4 cm and withing a degree, as shown in Table \ref{tab:idl}. 

\begin{sidewaystable*}[htbp]
\caption{Median and percentile localization errors for the Kitchen, Living Room and Baidu Indoor Localization Datasets. P stands for percentile, e.g., P 25\% means the maximum error for 25\% of the data when errors are sorted in the ascending order. m stands for metres}
  \centering
  
    \begin{tabular}{|l|l|c|c|c|c|c|}
    \specialrule{.15em}{.0em}{.0em} 
    \textbf{Indoor Datasets $\downarrow$} & \textbf{Errors $\downarrow$ \textbackslash{} Metrics $\rightarrow$} & \textbf{Median} & \textbf{P 25\%} & \textbf{P 50\%} & \textbf{P 75\%} & \textbf{P 90\%} \\\specialrule{.15em}{.0em}{.0em} 
    
    \multirow{2}[0]{*}{\textbf{Kitchen}} & \textbf{Position Error (m)} & 0.0117 m & 0.0066 m & 0.0117 m & 0.0223 m & 0.0395 m\\\cline{2-7}
          & \textbf{Angle Error (degrees)} & 0.2128\degree & 0.1448\degree & 0.2128\degree & 0.3845\degree  & 0.6399\degree \\\specialrule{.15em}{.0em}{.0em}
          
    \multirow{2}[0]{*}{\textbf{Living Room}} & \textbf{Position Error (m)} & 0.0101 m & 0.0059 m & 0.0101 m & 0.0180 m & 0.0470 m\\\cline{2-7}
          & \textbf{Angle Error (degrees)} & 0.3254\degree & 0.2011\degree & 0.3254\degree & 0.4877\degree & 1.6218\degree \\\specialrule{.15em}{.0em}{.0em}
          
    \multirow{2}[0]{*}{\textbf{Baidu Indoor Localization}} & \textbf{Position Error (m)} & 0.6930 m & 0.2020 m & 0.6930 m & 10.44 m & 24.55 m\\\cline{2-7}
          & \textbf{Angle Error (degrees)} & 2.30\degree & 0.6105\degree & 2.30\degree & 13.72\degree & 22.05\degree \\\specialrule{.15em}{.0em}{.0em}

    \end{tabular}
    \label{tab:idl}
\end{sidewaystable*}

\subsubsection{Living Room Dataset}
The Living Room Dataset is part of the Stanford RGB Localization Database \citep{valentin2016learning} and comprises the 3D model and 2D images of a living area in an apartment with a total volume of $30m^3$. This dataset was also captured with a combination of Structure.io sensor and iPad cameras. For the quantitative evaluation on this dataset, $20\%$ of the images were randomly held out for query set, while the remaining images were used for the training set. We were able to localize all the images in the query set with more than 90\% localizations within a 5 cm error (Table \ref{tab:idl}).

\begin{figure*}[htbp]
     \centering
     \begin{subfigure}[htbp]{1\textwidth}
         \centering
         \includegraphics[width=\textwidth]{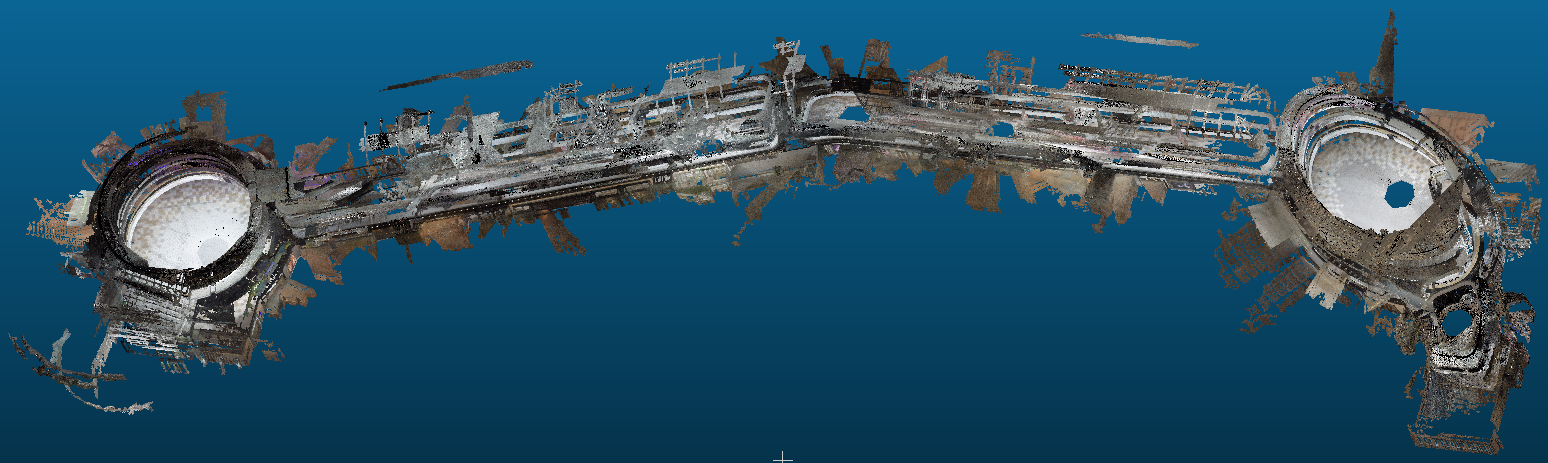}
         \caption{Top view}
         \label{fig:idl1}
     \end{subfigure}
     \hfill
     \begin{subfigure}[htbp]{1\textwidth}
         \centering
         \includegraphics[width=\textwidth]{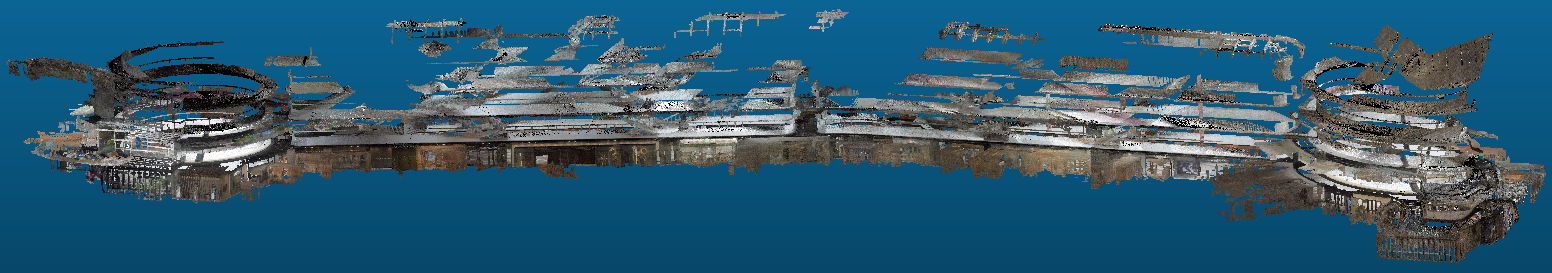}
         \caption{Side view}
         \label{fig:idl2}
     \end{subfigure}
        \caption{Different views of the point cloud of the shopping mall from Baidu Indoor Localization Dataset \citep{idl}. The large scale and repetitive structures in the dataset make it extremely challenging to localize 2D images in the point cloud compared to other indoor datasets. }
        \label{fig:idl}
\end{figure*}

\subsubsection{Baidu Indoor Localization Dataset}
 The Baidu Indoor Localization Dataset \citep{idl} is composed of a 3D point cloud of a multi-storey shopping mall created with the scans from a LIDAR scanner and a database of images with ground truth information for camera positions and orientations. The 3D point cloud contains more than 67 million points. Figure \ref{fig:idl} shows different views of the point cloud of the mall. The images in the dataset are captured with different cameras and smart phones and at different times which increases the complexity of the dataset. Moreover, many of the images contain occlusions due to the persons shopping in the mall. 
We randomly selected $10\%$ of the images for the query set, while the remaining images were used for the training set. Our technique successfully localized 78.2\% of the images in the query set. Despite the challenges of the dataset, we achieved a median pose estimation error of 0.69 m and 2.3 degrees as shown in Table \ref{tab:idl}.

\begin{sidewaystable*}[htbp]
\caption{Median errors for position and orientation estimation of our technique compared to other approaches on Cambridge Landmarks outdoor datasets. Pos stands for median positional error in meters and Ang stands for rotational error in degrees. NA: Results not available. Best results are shown in \textbf{bold} and second best results are \underline{underlined}.}
  \centering
 
    \begin{tabular}{|A|R|R|R|R|R|R|R|R|R|R|R|R|R|R|}
    \hline
   \textbf{Methods$\rightarrow$} & \multicolumn{2}{V|}{\textbf{PoseNet \citep{posenet} ICCV'15}} & \multicolumn{2}{V|}{\textbf{Geom. Loss Net \citep{kendall2017geometric} CVPR'17}} & \multicolumn{2}{V|}{\textbf{VLocNet \citep{valada2018deep} ICRA'18}} & \multicolumn{2}{V|}{\textbf{DSAC \citep{brachmann2017dsac} CVPR'17}} & \multicolumn{2}{V|}{\textbf{Active Search \citep{sattler2017efficient} TPAMI'17}} & \multicolumn{2}{V|}{\textbf{DSAC++ \citep{brachmann2018learning} CVPR'18}} &\multicolumn{2}{V|}{\textbf{Ours}} \\
    \hline
    \textbf{Methods' Type$\rightarrow$} & \multicolumn{2}{V|}{\textbf{Network-based}} & \multicolumn{2}{V|}{\textbf{Network-based}} & \multicolumn{2}{V|}{\textbf{Network-based}} & \multicolumn{2}{V|}{\textbf{Network + RANSAC}} & \multicolumn{2}{V|}{\textbf{SfM-based}} & \multicolumn{2}{V|}{\textbf{Network + RANSAC}} &\multicolumn{2}{V|}{\textbf{2D to 3D Descriptors Matching}} \\
    \hline
    \textbf{Datasets$\downarrow$} & \textbf{Pos (m)} & \textbf{Ang} & \textbf{Pos (m)} & \textbf{Ang} & \textbf{Pos (m)} & \textbf{Ang} & \textbf{Pos (m)} & \textbf{Ang} & \textbf{Pos (m)} & \textbf{Ang} & \textbf{Pos (m)} & \textbf{Ang} & \textbf{Pos (m)} & \textbf{Ang}\\
    \hline
    \textbf{Shop Facade} & 1.46 & 4.04\degree  & 0.88 & 3.78\degree   & 0.593 & 3.529\degree  & \underline{0.09} & \textbf{0.4\degree}    & 0.12 & \textbf{0.4\degree}    & \underline{0.09} & \textbf{0.4\degree} & \textbf{0.086} & \underline{0.779\degree}  \\
    \hline
    \textbf{Old Hospital} & 2.31 & 2.69\degree   & 3.2 & 3.29\degree    & 1.075 & 2.411\degree  & 0.33 & 0.6\degree    & 0.44 & 1\degree      & \underline{0.24} & \underline{0.5\degree} & \textbf{0.129} & \textbf{0.221\degree}  \\
    \hline
    \textbf{King's College} & 1.92 & 2.70\degree     & 0.88 & 1.04\degree     & 0.836 & 1.419\degree   & 0.30 & 0.5\degree    & 0.42  & 0.6\degree  &  \underline{0.23} & \underline{0.4\degree} & \textbf{0.087} & \textbf{0.147\degree}  \\
    \hline
    \textbf{Great Court } & NA & NA   & 6.83 & 3.47\degree   & NA & NA   & 2.80 & 1.5\degree    & NA & NA  & \underline{0.66} & \underline{0.4\degree} & \textbf{0.509} & \textbf{0.352\degree}  \\
    \hline
    \textbf{St. Mary Church} & 2.65 & 4.24\degree   & 1.57 & 3.32\degree    & 0.631 & 3.906\degree   & 0.55 & 1.6\degree      & \underline{0.19} & \underline{0.5\degree} & 0.20 & 0.7\degree  & \textbf{0.147} & \textbf{0.467\degree}  \\
    \hline
    \end{tabular}
  \label{tab:comparison}
\end{sidewaystable*}
\subsection{Comparison with Other Approaches}
The proposed technique for image localization in point clouds is based on a novel concept of directly matching the descriptors extracted from images and point clouds. The SfM based techniques are based on the matching of 2D features and require the point cloud to be generated from SfM pipeline. On the other hand, our technique can work with point clouds created with any 3D scanner or generated with any method and has no dependency on any information from SfM. 

The network based regression methods train directly to regress the poses of the images. However, this causes the networks to over-fit on the ground truth poses. Therefore at test time, they only produce good results for the images with poses close to the training ones \citep{sattler2019understanding}. In our technique, training of the Descriptor-Matcher is carried out on the feature descriptors with no information of the poses or the location of the points. This ensures that the Descriptor-Matcher does not over-fit on the training poses, rather it learns to find a mapping between the 2D and 3D descriptors. The final pose estimation is based on the principles of geometry which produce better pose estimates compared to end-to-end trained methods \citep{sattler2019understanding}. Therefore, the proposed method benefits from the non-reliance on SfM pipeline like the network based methods, as well as the ability to use geometry based pose estimation similar to the SfM based methods.

For quantitative analysis, we provide a comparison of our results with the 
state-of-the-art methods
for camera pose localization on the common datasets between the various approaches. We compare the median values of errors in the estimation of camera position and rotation on the outdoor datasets of Cambridge Landmarks Database \citep{posenet}, as mostly only the median errors are reported by other methods. The results of the compared methods are not available for Baidu Indoor Localization Dataset or Stanford RGB Localization Database.
Our proposed method achieved competitive or superior localization accuracy to the state-of-the-art methods for 6-DOF pose estimation on all the datasets. 
Table \ref{tab:comparison} shows the median values for position and rotational errors of our technique compared to other prominent methods.

\section{Conclusion}\label{conclusion}
This paper proposed a novel method to directly match descriptors from 2D images with those from 3D point clouds, where the descriptors are extracted using 2D and 3D feature extraction techniques, respectively. 
We have shown that direct matching of 2D and 3D feature descriptors is an unconstrained and reliable method for 6-DOF pose estimation in point clouds. 
Our approach results in a pipeline which is much simpler compared to SfM or network based methods. 
Extensive quantitative evaluation has demonstrated that the proposed method achieved competitive performance with the state-of-the-art methods in the field.
Moreover, unlike SfM based techniques, the proposed method can be used with point clouds generated with any type of 3D scanners and can work in both indoor and outdoor scenarios.

\section*{Acknowledgements}

This work was supported by the SIRF scholarship from the University of Western Australia (UWA) and by the Australian Research Council under Grant DP150100294.

\balance

\bibliography{Unconstrained_descriptor_matching}
\end{document}